\documentclass[runningheads]{llncs}

\usepackage{eccv}

\usepackage{eccvabbrv}

\usepackage{graphicx}
\usepackage{booktabs}
\usepackage{multirow}
\usepackage{bm}
\usepackage{algorithm}
\usepackage{algorithmic}

\usepackage[accsupp]{axessibility}

\usepackage[hidelinks]{hyperref}
\usepackage{orcidlink}

\newcommand{\R}{\mathbb{R}}
\newcommand{\ours}{GridFlow}

\begin{document}

\title{GridFlow: Structured Latent Flow for Seamless City-Scale 3D Point Cloud Generation}

\titlerunning{GridFlow}

\author{Xinyu~Wang\orcidlink{0009-0008-4065-2472} \and
Muhammad~Ibrahim\orcidlink{0000-0002-5376-2477} \and
Atif~Mansoor\orcidlink{0000-0001-6940-3914} \and
Ajmal~Mian\orcidlink{0000-0002-5206-3842}}

\authorrunning{X.~Wang et al.}

\institute{The University of Western Australia, Perth, WA 6009, Australia \\
\email{ajmal.mian@uwa.edu.au}}

\maketitle

\begin{abstract}
Generating realistic 3D city environments from remote sensing data is important for simulation, urban planning, and mixed reality, yet existing point cloud generation methods are limited to single objects or bounded indoor scenes and cannot handle the scale, seamless tiling, and partial observability challenges of city-scale generation.
We present \ours{}, a multi-stage framework that generates dense, colored point clouds ($10^5$ points per $150\text{m}{\times}150\text{m}$ tile) at city scale, conditioned on satellite imagery, semantic segmentation maps, and digital surface models (DSM).
A \emph{Grid-Aligned VAE} encodes each tile into a topology-preserving latent grid where tokens correspond to fixed spatial regions, enabling spatially coherent multi-modal conditioning and compact latent-space edge consistency that implicitly aligns thousands of boundary points for seamless cross-tile generation.
A conditional rectified flow model synthesizes geometry latents from the fused multi-modal conditions, and an orientation-aware diffusion colorizer separately handles satellite-visible horizontal surfaces and occluded vertical fa\c{c}ades.
To support standardized evaluation, we build on public 3D data sources to introduce \emph{City3D-MultiGen}, a benchmark of $163$K densely annotated tiles from Melbourne and London with aligned point clouds, satellite images, semantic maps, and elevation data.
Experiments show that \ours{} outperforms adapted point cloud generation baselines across all geometry metrics and produces visually coherent colored point clouds with seamless boundaries over arbitrarily large urban extents. Our benchmark details are available at \url{https://huggingface.co/datasets/e32/City3D-MultiGen}.
\end{abstract}

\section{Introduction}
\label{sec:intro}

\begin{figure}
    \centering
    \includegraphics[width=1\linewidth]{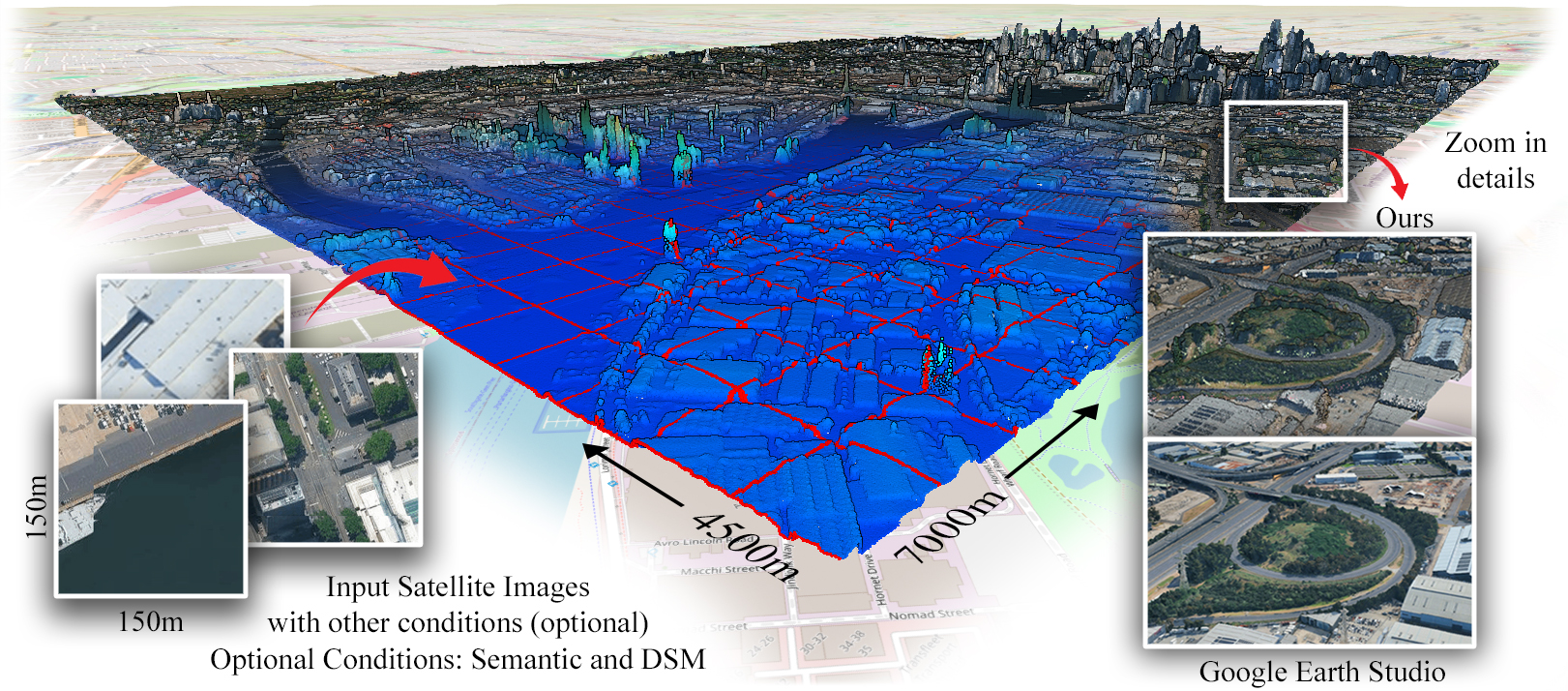}
    \caption{Overview of \ours{} visualized in Potree~\cite{schuetz2016potree}. Given a satellite image and optional semantic/DSM conditions, our three-stage pipeline generates dense, colored city-scale point clouds with seamless cross-tile boundaries.}
    \label{fig:overview}
\end{figure}

Realistic 3D city environments are central to autonomous driving simulation, urban planning, and virtual content creation.
While photogrammetric reconstruction yields accurate geometry from dense multi-view images, its dependence on extensive aerial capture limits scalability and prevents content creation for hypothetical or data-scarce regions.
Generative models offer a compelling alternative: synthesizing plausible 3D city geometry directly from widely available remote sensing data such as satellite images (see Fig.~\ref{fig:overview}).

Generating 3D environments has seen growing interest across multiple representation paradigms.
Point cloud generation methods, including normalizing flows~\cite{yang2019pointflow}, denoising diffusion~\cite{luo2021diffusion,zhou2021pvd}, score-based models~\cite{cai2020shapegf}, latent diffusion~\cite{zeng2022lion}, diffusion transformers~\cite{mo2023dit3d}, and text-conditioned systems~\cite{nichol2023pointe}, have demonstrated impressive fidelity on object benchmarks such as ShapeNet~\cite{chang2015shapenet,du2025superpc} and have been extended to indoor scene synthesis~\cite{hollein2023text2room,zhang2023commonscenes,tang2024diffuscene,wu2024blockfusion}. However, these approaches typically operate on bounded scenes and do not address city-scale tiling, multi-modal remote sensing conditioning, or seamless boundary consistency.

Scaling to outdoor scenes, recent works have explored semantic voxel diffusion for driving scenarios~\cite{lee2024semcity,liu2024pyramid}, sparse voxel hierarchies for $100\text{m}$-scale generation~\cite{ren2024xcube}, LiDAR point cloud diffusion for scene completion~\cite{nunes2024lidiff,ran2024lidm}, and semantic-aware range-view generation~\cite{zhu2025spiral}.
In the city-generation domain, neural-field-based approaches~\cite{xie2024citydreamer,chen2023scenedreamer,xie2025citydreamer4d} and Gaussian splatting methods~\cite{xie2025gaussiancity,lee2025skyfallgs} produce visually compelling renderings but lack explicit geometry.
Despite this progress, generating dense, colored point clouds at city scale from remote sensing data remains open, with three core challenges:
(i)~\emph{Scale}---a single $150\text{m}{\times}150\text{m}$ tile contains ${\sim}10^5$ points, making direct generation intractable; voxel methods~\cite{ren2024xcube,lee2024semcity} sacrifice detail, and range-view approaches~\cite{ran2024lidm,zhu2025spiral} couple density to sensor resolution.
(ii)~\emph{Seamless tiling}---independently generated tiles must join without seams, yet matching millions of boundary points is prohibitively expensive; existing methods either lack tiling~\cite{lee2024semcity,liu2024pyramid} or resort to post-hoc blending~\cite{lin2023infinicity}.
(iii)~\emph{Partial observability}---satellite sensors see horizontal surfaces but not vertical fa\c{c}ades, requiring complex and separate colorization strategies.

We address these challenges with \ours{}, a multi-stage framework built around a structured latent representation (Fig.~\ref{fig:pipeline}).
A \emph{Grid-Aligned VAE} (GA-VAE) maps each city tile to a structured latent grid $\mathbf{z}\in\R^{64\times25\times25}$ where each token corresponds to a fixed $6\text{m}{\times}6\text{m}$ physical region.
This spatial grounding serves two purposes: it aligns the latent space with the condition modalities for spatially coherent conditioning, and reduces generation from a $10^5$-point output space to a compact 625-token latent space.
Since the latent grid is topology-preserving, a simple MSE loss on the boundary columns of adjacent tile latents implicitly aligns thousands of decoded boundary points, achieving seamless city-scale tiling at negligible additional cost.
A conditional rectified flow model then synthesizes latent codes from fused satellite, semantic, and elevation features injected at spatially matching grid positions, while an orientation-aware diffusion colorizer handles the observability asymmetry by applying satellite-guided attention to horizontal surfaces and synthesizing vertical textures from a learned fa\c{c}ade pattern bank.
To support standardized evaluation, we construct \emph{City3D-MultiGen}, a benchmark of $163$K densely annotated tiles built on top of two publicly available 3D data sources, the City of Melbourne 3D Point Cloud~\cite{melbourne2018pointcloud} and HoliCity~\cite{zhou2020holicity}, covering Melbourne (Australia) and London (UK) respectively, with aligned multi-modal data. Our contributions include:

\begin{enumerate}
  \item A Grid-Aligned VAE that maps city scenes to a spatially grounded 2D latent grid with explicit token-to-region correspondence, enabling cross-modal conditioning and seamless boundary-aware generation.

  \item A latent-space edge consistency mechanism that constrains adjacent tile latents rather than decoded point clouds, reducing the matching problem by two orders of magnitude in dimensionality and enabling seamless infinite-extent city-scale generation.

  \item An orientation-aware diffusion colorization model that addresses the partial observability of satellite imagery through dual-path conditioning, separately handling directly visible horizontal surfaces and occluded vertical fa\c{c}ades.

  \item A large-scale benchmark of $163$K tiles from two cities with aligned point clouds, satellite images, semantic segmentation, and DSM, together with a fully automated processing pipeline for reproducible evaluation of conditional city-scale 3D generation.
\end{enumerate}

\section{Related Work}
\label{sec:related}

\subsubsection{Object-Level Point Cloud Generation.}
Early deep generative models for point clouds operate on single objects from benchmarks like ShapeNet~\cite{chang2015shapenet}.
Achlioptas~\etal~\cite{achlioptas2018learning} introduced auto-encoder-based latent-space GANs, followed by PointFlow~\cite{yang2019pointflow} which models continuous normalizing flows, DPM~\cite{luo2021diffusion} which applies denoising diffusion directly in point space, and LION~\cite{zeng2022lion} which performs latent diffusion with a hierarchical VAE.
These methods produce high-quality shapes of 2--4K points but assume centered, single-object inputs and offer no mechanism for spatial conditioning, cross-tile consistency, or scaling beyond ${\sim}10^4$ points.

\subsubsection{Indoor Scene Generation.}
Beyond single objects, several methods tackle room-scale 3D scene synthesis.
Text2Room~\cite{hollein2023text2room} generates textured room meshes from text prompts by iteratively inpainting and fusing depth maps.
CommonScenes~\cite{zhang2023commonscenes} produces layout-conditioned 3D indoor scenes guided by scene graphs, while DiffuScene~\cite{tang2024diffuscene} formulates indoor layout generation as a denoising diffusion process over object arrangements.
Wu~\etal~\cite{wu2023sketch} introduce sketch-and-text guided diffusion for colored point cloud generation, later extended to leverage external knowledge for richer 3D scene synthesis from sketches~\cite{wu2024external}.
While these methods advance scene-level generation, they target bounded indoor environments and do not address the tiling, scale, or remote sensing conditioning requirements of city-scale generation.

\subsubsection{Outdoor Scene-Level Generation.}
Recent work extends diffusion models to outdoor scenes at larger scales.
XCube~\cite{ren2024xcube} generates sparse voxel hierarchies up to $1024^3$ resolution for $100\text{m}{\times}100\text{m}$ outdoor scenes, while SemCity~\cite{lee2024semcity} and PDD~\cite{liu2024pyramid} produce semantic occupancy grids via triplane and coarse-to-fine discrete diffusion, respectively.
LiDiff~\cite{nunes2024lidiff} and LiDM~\cite{ran2024lidm} further apply point- and range-image-level diffusion for scene completion and generation, Spiral~\cite{zhu2025spiral} jointly produces depth, reflectance, and semantic labels via a unified range-view diffusion model, Skip Mamba Diffusion~\cite{liang2025skipmamba} combines state-space models with diffusion for monocular 3D semantic scene completion, UrbanDiff~\cite{zhang2024urbandiff} and DynamicCity~\cite{bian2025dynamiccity} generate semantic occupancy from driving data, and LT3SD~\cite{meng2025lt3sd} enables patch-by-patch infinite 3D scene generation via latent tree diffusion.
These methods operate primarily on driving-centric data (SemanticKITTI, nuScenes) and are either unconditional or conditioned on LiDAR scans rather than satellite imagery.

\subsubsection{City-Scale 3D Generation.}
SceneDreamer~\cite{chen2023scenedreamer}, CityDreamer~\cite{xie2024citydreamer}, and CityDream\-er4D~\cite{xie2025citydreamer4d} produce unbounded city scenes via neural hash-grid fields but store geometry only implicitly.
InfiniCity~\cite{lin2023infinicity} tiles octree voxels with neural rendering. GaussianCity~\cite{xie2025gaussiancity}, Skyfall-GS~\cite{lee2025skyfallgs}, and Horizon-GS~\cite{jiang2025horizongs} adopt Gaussian splatting for city scenes but produce Gaussian primitives rather than geometric point clouds.
ImpliCity~\cite{stucker2022implicity} learns implicit occupancy fields from satellite images for city-scale surface reconstruction, but targets reconstruction rather than generation.
Closer to our setting, Sat2City~\cite{hua2025sat2city}, MagicCity~\cite{yao2025magiccity}, and Sat2RealCity~\cite{kang2025sat2realcity} generate 3D city content from satellite imagery via diffusion or multi-view synthesis, and CymbaDiff~\cite{liang2025cymbadiff} generates semantic city scenes from sketches.

A parallel line of work targets unbounded scene generation by tiling object- or chunk-level generators. NuiScene~\cite{lee2025nuiscene} trains an explicit outpainting model, while SynCity~\cite{engstler2025syncity}, TRELLISWorld~\cite{chen2025trellisworld}, and Extend3D~\cite{yoon2026extend3d} compose overlapping tiles via 2D inpainting, cosine-weighted latent blending, or coupled denoising, respectively. These methods, along with the block-wise extrapolation of BlockFusion~\cite{wu2024blockfusion}, achieve cross-tile coherence at inference time through overlapping regions, blending, or joint denoising, which couples generation cost to the overlap and can leave residual seams that require post-hoc registration. In contrast, our latent-space edge consistency is a learned constraint imposed during training, allowing tiles to be generated fully independently at inference without overlap or blending.

Our work differs in three respects: (i)~we condition on widely available satellite images and semantic maps rather than requiring 3D sensor inputs, (ii)~we generate explicit colored point clouds instead of voxel occupancy, and (iii)~we enforce latent-space edge consistency for seamless tiling, a requirement absent in prior city-scale or driving-scenario methods.

\section{City3D-MultiGen Benchmark}
\label{sec:datasets}

\begin{figure}[t]
\centering
\includegraphics[width=\textwidth]{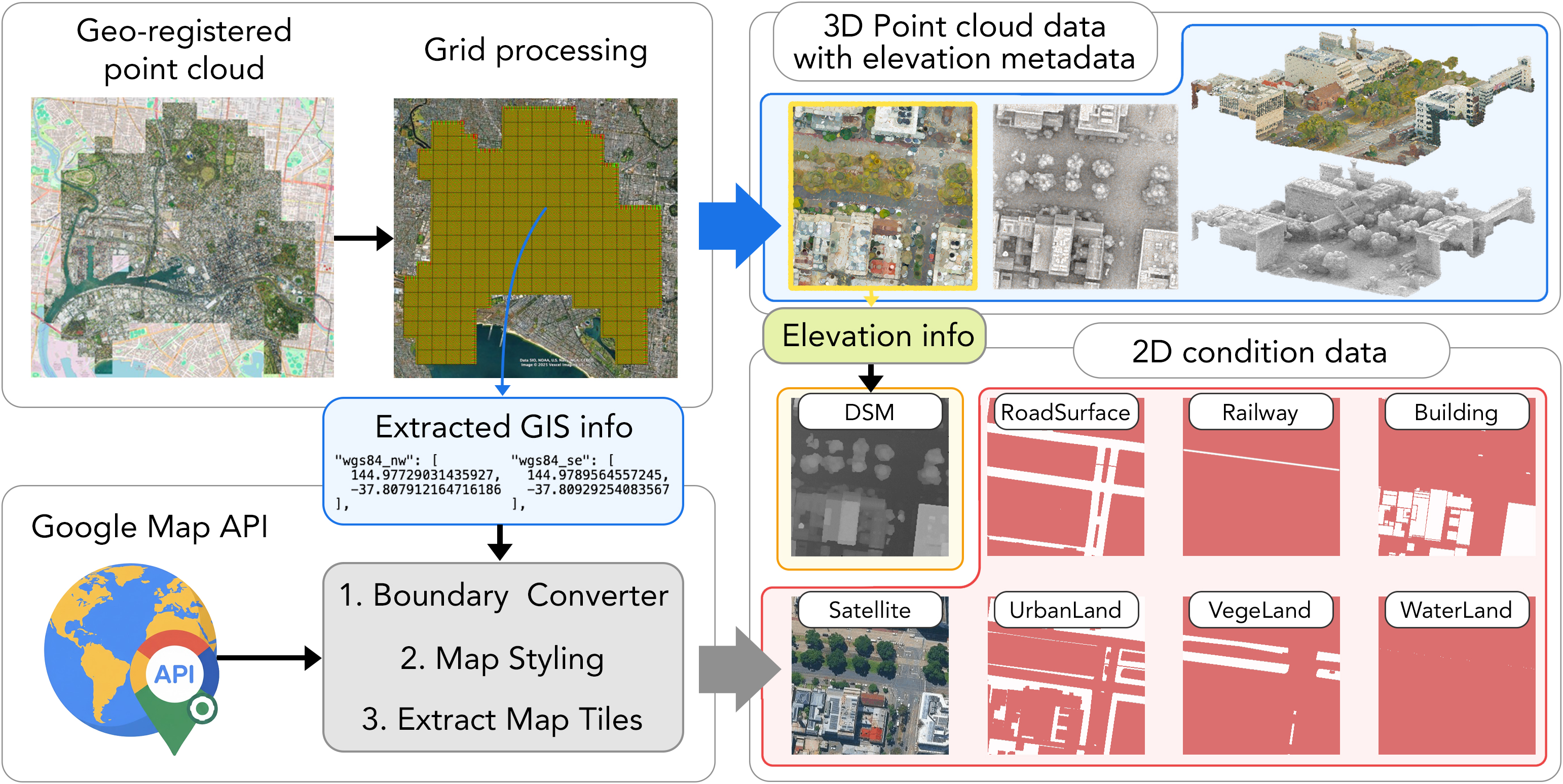}
\caption{Overview of the City3D-MultiGen data processing pipeline. Geo-registered point clouds are partitioned into $150\text{m}{\times}150\text{m}$ tiles on a regular grid. Geographic coordinates extracted from each tile are used to query the Google Maps API, which retrieves spatially aligned satellite imagery, styled semantic segmentation maps (via custom map styling), and DSM rasterized from the point cloud elevation. The resulting multi-modal tile set provides paired 3D geometry and multi-modal condition data for training and evaluation.}
\label{fig:dataset}
\end{figure}

Existing city-scale 3D generation methods typically evaluate on small-scale or synthetic scenes, lacking large-scale benchmarking with aligned multi-modal conditions. To address this gap, we construct \emph{City3D-MultiGen}, a benchmark comprising 163K densely annotated tiles from two cities on different continents, each with accurately aligned point cloud geometry, satellite images, semantic segmentation, and elevation data. We release the full processing pipeline code, tile metadata, and ready-to-run data retrieval scripts that allow users to reproduce the complete dataset at \url{https://huggingface.co/datasets/e32/City3D-MultiGen}. The point clouds are obtained from their original public providers, and the satellite imagery and semantic maps through these scripts, rather than redistributed directly, as the source-data licenses and the map provider's terms of service do not permit redistribution. An overview of the data processing pipeline is shown in Fig.~\ref{fig:dataset}.

\subsubsection{Source Data.}
We draw on two complementary 3D sources:
(i)~The City of Melbourne 3D Point Cloud~\cite{melbourne2018pointcloud} covers ${\sim}37.7$\,km$^2$ of metropolitan Melbourne, Australia, captured via aerial photogrammetry at 7.5\,cm ground sample distance with 25\,cm spatial accuracy. The dense, colorized point cloud is distributed as georeferenced LAS tiles with RGB information.
(ii)~HoliCity~\cite{zhou2020holicity} provides high-fidelity publicly accessible FBX models covering 2\,km$^2$ of central London, United Kingdom. Point clouds were extracted by uniformly sampling the mesh surfaces of the FBX model files.
The two city datasets differ substantially: Melbourne is largely suburban with low-rise buildings, while London is dense and structurally complex, enabling cross-domain evaluation.

\subsubsection{Automated Multi-Modal Alignment Pipeline.}
Both datasets are processed through an identical, fully automated pipeline with no manual annotation. The geographic extent is partitioned into $150\text{m}{\times}150\text{m}$ tiles on a regular UTM grid with 20\,m center spacing. Each tile is paired with three spatially aligned condition modalities at $256{\times}256$ resolution: (i)~\textbf{satellite imagery} (RGB) retrieved via the Google Maps Static API, (ii)~\textbf{semantic segmentation} (6 binary class masks: \emph{Building}, \emph{RoadSurface}, \emph{Railway}, \emph{VegetationLand}, \emph{UrbanLand}, \emph{WaterSurface}) obtained by parsing custom-styled Google Maps renders, and (iii)~a \textbf{DSM} rasterized from point cloud elevations and normalized to $[0,1]$.
Point coordinates are centered per tile and normalized to $[-1,1]^3$, with the vertical axis scaled independently because building heights (typically 10--50\,m) are much smaller than the $150\text{m}$ horizontal tile extent; uniform scaling would compress height variations into a narrow range, losing fine vertical detail. Each raw tile contains approximately $150{,}000$ points.
The dataset is split 80/10/10 into training, validation, and test sets based on contiguous spatial regions along the grid coordinate axes, with a minimum separation of ${\geq}150\text{m}$ between splits, ensuring no geographic overlap between training and test data.

\section{Method}
\label{sec:method}

The primary input to our method is a single satellite image of a $150\text{m}{\times}150\text{m}$ city tile at $256{\times}256$ resolution. Semantic segmentation maps (six land-cover classes), which can be extracted automatically from the satellite image, and DSM, when available, serve as auxiliary conditions that further refine spatial layout and elevation. Structured condition dropout during training (Sec.~\ref{sec:flow}) ensures that the model remains effective under any subset of available modalities. The output of our method is a colored point cloud of $N{=}100{,}000$ points.
To the best of our knowledge, this is the first method capable of generating a seamless city-scale point cloud of this magnitude.

\begin{figure}[t]
    \centering
    \includegraphics[width=1\linewidth]{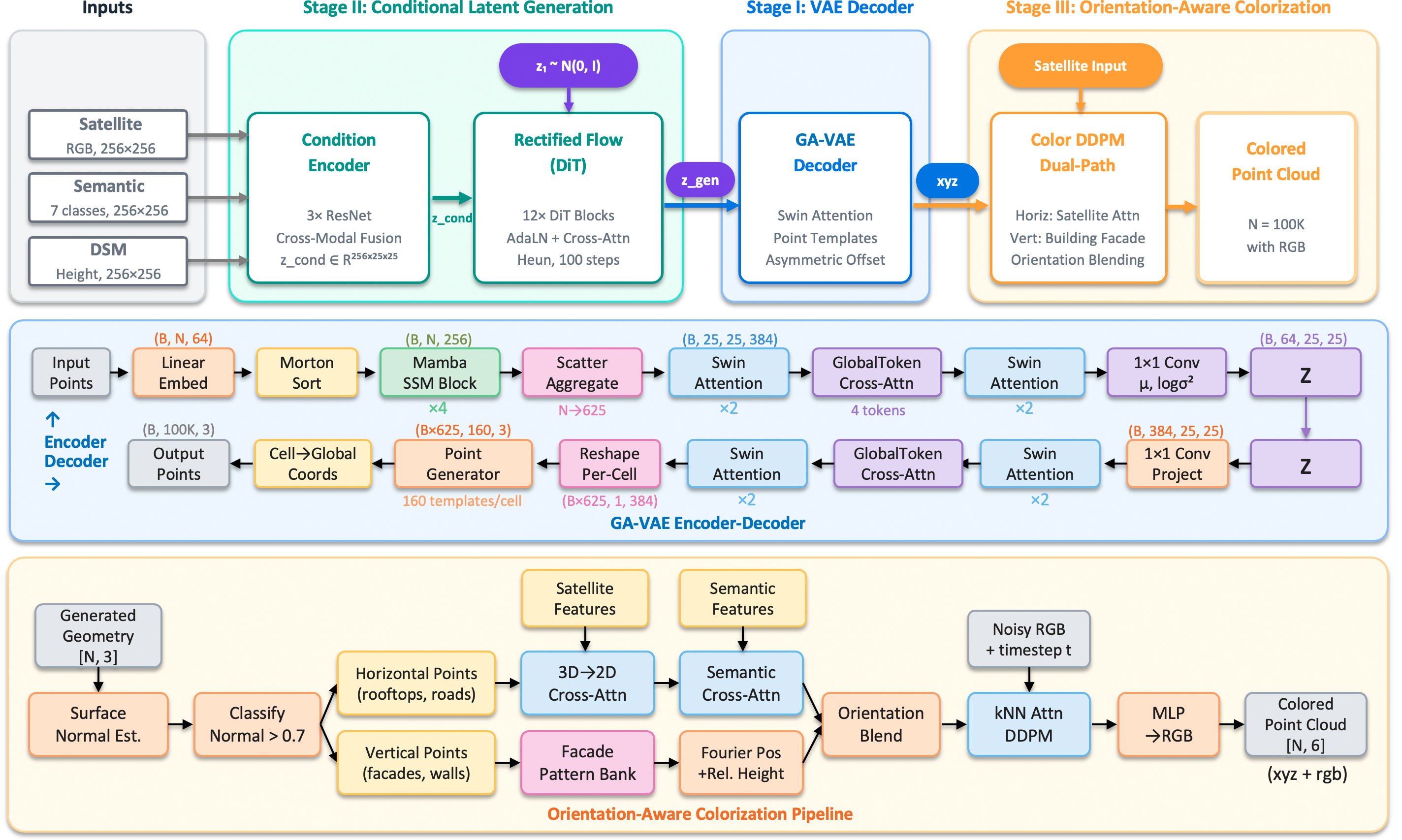}
        \caption{Overview of the proposed \ours{} framework. \textbf{Stage~I}: A Grid-Aligned VAE compresses 100K-point city clouds into a topology-preserving $64{\times}25{\times}25$ latent grid. \textbf{Stage~II}: A conditional latent generator synthesizes geometry latents with boundary consistency constraints from fused satellite/semantic/DSM features. \textbf{Stage~III}: An orientation-aware colorization model produces realistic appearance via dual-path surface conditioning.}
    \label{fig:pipeline}
\end{figure}

Our method comprises three stages as shown in Fig.~\ref{fig:pipeline}: (1)~a \emph{Grid-Aligned Variational Autoencoder} (GA-VAE) that learns a topology-preserving latent representation of city-scale point clouds (Sec.~\ref{sec:vae}), (2)~a \emph{conditional latent generator} that synthesizes geometry in the learned latent space from the fused input modalities (Sec.~\ref{sec:flow}), and (3)~an \emph{orientation-aware colorization} model that produces realistic appearance for the generated geometry (Sec.~\ref{sec:color}). Our design avoids the training instability of joint geometry-color optimization and allows each stage to employ the most suitable generative framework.

\subsection{Grid-Aligned Variational Autoencoder}
\label{sec:vae}

City environments exhibit natural spatial regularity. Buildings, roads, and vegetation form locally coherent clusters throughout the horizontal plane. The GA-VAE uses this regularity by dividing the horizontal plane into a $25{\times}25$ grid of $6\text{m}{\times}6\text{m}$ cells. Each cell is encoded separately, and the model then combines information across cells in a hierarchical way. The latent $\mathbf{z} \in \R^{64\times25\times25}$ keeps a one-to-one link between tokens and physical regions to facilitate spatially aligned conditioning (Sec.~\ref{sec:condition}) and latent-space edge consistency (Sec.~\ref{sec:edge}).

\subsubsection{Encoding.}
Each cell contains ${\sim}160$ points on average. Points are sorted along a Morton curve~\cite{morton1966computer} and encoded by stacked selective state-space blocks (SSM)~\cite{gu2023mamba} in $\mathcal{O}(N)$ time into a fixed-dimensional cell descriptor. The 625 cell features are then arranged on the $25{\times}25$ grid and refined by Swin~\cite{liu2021swin} windowed self-attention layers for local communication, augmented with a bottleneck cross-attention module at the mid-layer: a small set of learnable global tokens (4 tokens) aggregates scene-level information from all cells and broadcasts it back, enabling global context without full attention. The same module is applied symmetrically in the decoder.

\subsubsection{Decoding.}
A symmetric windowed-attention decoder maps the latent $\mathbf{z}$ back to per-cell features. For each cell, \emph{learnable point templates} ($160$ points initialized from $\mathcal{N}(0, 0.09)$) are refined via cross-attention to the decoded cell feature, followed by self-attention among the template points. Separate MLP heads predict horizontal and vertical offsets with \emph{asymmetric scaling}: horizontal offsets are scaled by $0.5$ (matching the $6\text{m}$ cell extent), while vertical offsets are scaled by $4.0$ (accommodating the much larger range of building heights), with a learned per-cell height bias anchoring the generated points to the appropriate elevation. The final coordinates are obtained by adding offsets to cell centers, yielding $25^2 \times 160 = 100{,}000$ points.

\subsubsection{Training Objective.} 
The training objective is defined as
\begin{equation}
  \mathcal{L}_\text{VAE} = \mathcal{L}_\text{CD} + \lambda_h \mathcal{L}_\text{height} + \lambda_b \mathcal{L}_\text{bnd} + \lambda_\text{KL} \mathcal{L}_\text{KL},
  \label{eq:vae_loss}
\end{equation}
where $\mathcal{L}_\text{CD}$ is the bidirectional Chamfer distance, $\mathcal{L}_\text{height}$ is a KL divergence between soft height histograms that regularizes vertical structure, $\mathcal{L}_\text{bnd}$ is Chamfer distance on tile-boundary points to preserve reconstruction fidelity near tile edges, and $\mathcal{L}_\text{KL}$ is the standard VAE regularization with a very low weight ($\lambda_\text{KL}{=}10^{-5}$) to prioritize reconstruction.

\subsection{Conditional Latent Generation}
\label{sec:flow}

\subsubsection{Multi-Modal Condition Fusion.}
\label{sec:condition}
Three ResNet-style encoders extract features from satellite images, semantic segmentation maps, and DSM respectively, with capacity proportional to each modality's information density. The features are projected to a common dimensionality and fused via bidirectional cross-modal attention, where each modality queries the other two. The fused representation $\mathbf{z}_\text{cond} \in \R^{256\times25\times25}$ is spatially aligned with the VAE latent grid. Structured condition dropout randomly zeros subsets of modalities during training, so the model learns to generate both conditionally and unconditionally; at inference, classifier-free guidance~\cite{ho2022classifier} extrapolates beyond the conditional prediction by contrasting it with the unconditional one, sharpening output fidelity. This also ensures robustness when some modalities are unavailable.

\subsubsection{Rectified Flow.}
Geometry latents are generated via rectified flow matching~\cite{liu2023flow,lipman2023flow}, which learns a velocity field transporting Gaussian noise to data along straight paths. Before flow training, VAE latents are normalized to zero mean and unit variance per channel, stabilizing the velocity targets. Given noise $\mathbf{z}_1 \sim \mathcal{N}(\mathbf{0}, \mathbf{I})$ and a normalized VAE latent $\mathbf{z}_0$, the interpolant is $\mathbf{z}_t = (1{-}t)\mathbf{z}_0 + t\mathbf{z}_1$ with target velocity $\mathbf{v} = \mathbf{z}_1 - \mathbf{z}_0$. A Transformer-based velocity network~\cite{peebles2023scalable} operating on the flattened $25{\times}25$ latent tokens is trained with:
\begin{equation}
  \mathcal{L}_\text{flow} = \mathbb{E}_{\mathbf{z}_0, \mathbf{z}_1, t}\!\left[\|\mathbf{v}_\theta(\mathbf{z}_t, t, \mathbf{z}_\text{cond}) - (\mathbf{z}_1 - \mathbf{z}_0)\|^2\right].
  \label{eq:flow_loss}
\end{equation}
Conditions enter via two paths: a \emph{global path} that modulates each transformer layer through AdaLN scale/shift parameters derived from the pooled condition and timestep, and a \emph{local path} where $25{\times}25$ condition tokens serve as cross-attention keys/values every two blocks, gated by a learned sigmoid. At inference, sampling uses 50-step Heun integration with classifier-free guidance~\cite{ho2022classifier}.

\subsubsection{Latent-Space Edge Consistency.}
\label{sec:edge}
City-scale generation requires seamless tiling across independently processed $150\text{m}$ tiles. Rather than matching $10^5$ boundary points in the output space, consistency is enforced directly in the compact latent space. For adjacent tiles $T_a, T_b$, the boundary columns (or rows) of their latents are constrained to agree:
\begin{equation}
  \mathcal{L}_\text{edge} = \frac{1}{2CW_e}\sum_{c,h}\sum_{w=1}^{W_e}\!\big(\mathbf{z}^{(a)}_{c,h,W-w} - \mathbf{z}^{(b)}_{c,h,w-1}\big)^2,
  \label{eq:edge_loss}
\end{equation}
where $W_e{=}2$ is the boundary margin width (${\sim}12\text{m}$), with an analogous formulation for vertical neighbors. During training, each sample draws an adjacent tile with probability $0.5$. Because the latent grid is topology-preserving, matching two columns of 64-dimensional tokens implicitly aligns the geometry of ${\sim}6{,}400$ boundary points, making this constraint both efficient and effective.

\subsection{Orientation-Aware Colorization}
\label{sec:color}

The observability asymmetry between horizontal and vertical surfaces demands fundamentally different colorization strategies. Rooftops and roads are directly visible in satellite images and can be colored via learned 3D-to-2D attention; building fa\c{c}ades, however, are invisible to top-down sensors and must instead be synthesized from learned texture priors. Prior point cloud colorization methods~\cite{shinohara2021point2color,gao2023sgnet} assume full visibility; our satellite setting requires handling partial observability, motivating a dual-path DDPM~\cite{ho2020denoising} architecture.

\subsubsection{Dual-Path Conditioning.}
A point is labeled horizontal when the vertical component of its normal is greater than 0.7, and vertical otherwise. The \emph{horizontal path} uses cross-attention between 3D point coordinates and satellite/semantic/DSM features, learning an implicit 3D-to-2D correspondence that accommodates varying viewing angles without requiring orthorectification. The \emph{vertical path} employs a learned bank of canonical fa\c{c}ade texture patterns, selected and modulated by multi-scale Fourier positional features, relative height within each building, and semantic class embeddings. An orientation-aware blending network produces per-point weights that smoothly combine both paths, mixing learned and geometric priors.

\subsubsection{Diffusion and Training.}
The fused conditioning features and noisy RGB tokens are processed by local $k$-NN attention blocks with relative positional encoding. Training proceeds in three phases that progressively introduce loss objectives: basic noise prediction and $L_1$ reconstruction (with $2{\times}$ weight on vertical surfaces), then $k$-NN smoothness regularization for local color coherence, and finally a vertical texture gradient loss that enforces floor-level repetition patterns characteristic of building fa\c{c}ades. This curriculum prevents the model from collapsing to averaged appearances before learning fine-grained structures.

\subsection{Implementation Details}
\label{sec:training}

The three stages are trained sequentially on the City of Melbourne 3D Point Cloud~\cite{melbourne2018pointcloud}, each freezing all upstream models: the VAE for 150 epochs, the flow model for 200 epochs, and the colorization model for 50 epochs. AdamW with cosine learning rate decay is used throughout. The color model trains on VAE-reconstructed (rather than ground-truth) geometry with latent noise augmentation to improve robustness to imperfect flow-generated latents at inference. Training the full pipeline takes approximately 55 days on a single NVIDIA RTX 5090 GPU (29/14/12 days for the GA-VAE, flow, and colorization stages, respectively). At inference, \ours{} generates one $150\text{m}{\times}150\text{m}$ tile in ${\sim}4.9$\,s and a $1\text{km}^2$ area (${\sim}49$ tiles) in ${\sim}4$\,minutes.

\section{Experiments}
\label{sec:experiments}

\subsubsection{Setup.}
We train \ours{} on the Melbourne split of City3D-MultiGen and evaluate on a held-out test set of 13{,}363 tiles. Following the evaluation protocol of Achlioptas~\etal~\cite{achlioptas2018learning}, we report Chamfer Distance (CD), F-Score ($\tau{=}0.05$), Jensen--Shannon Divergence (JSD) on a $28^3$ voxel grid, Coverage (COV-CD), and 1-Nearest-Neighbor Accuracy (1-NNA-CD). All metrics are computed on 500 randomly sampled test tiles.
%
Given the lack of city-scale point cloud generation methods, we adapt four baselines to our setting for comparison. These include:
PointFlow~\cite{yang2019pointflow}, DPM~\cite{luo2021diffusion}, PVD~\cite{zhou2021pvd}, and SemCity~\cite{lee2024semcity}. For visual rendering evaluation, we additionally compare with CityDreamer~\cite{xie2024citydreamer} and Skyfall-GS~\cite{lee2025skyfallgs}, using PSNR, SSIM, LPIPS, FID, and histogram correlation against Google Earth reference views.

\subsubsection{Geometry Comparison.}

\begin{figure}[t]
    \centering
    \includegraphics[width=1\linewidth]{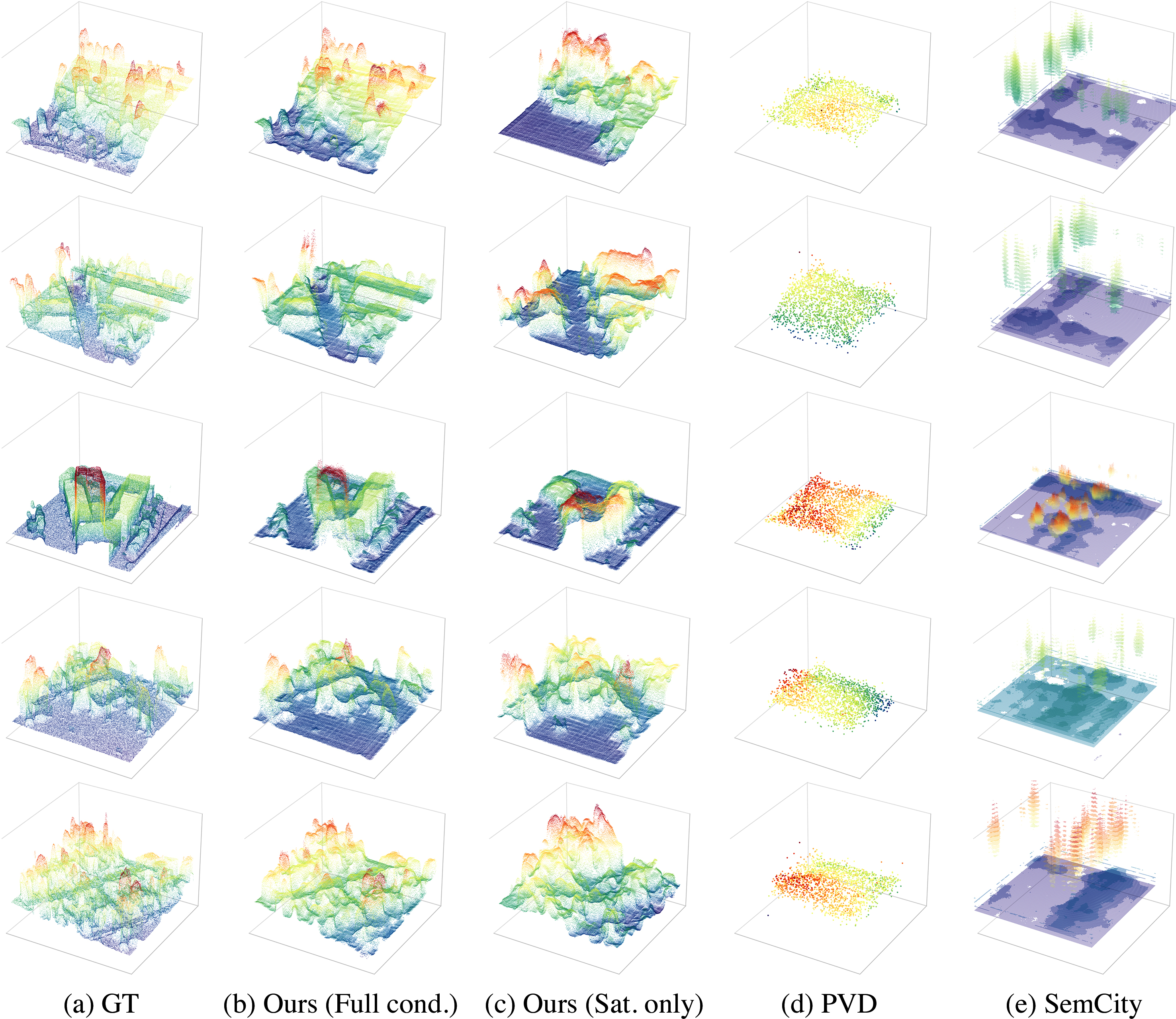}
    \caption{Qualitative geometry comparison on Melbourne test tiles. \ours{} produces detailed building structures and sharp boundaries compared with other methods.}
    \label{fig:geo_compare}
\end{figure}

\begin{table}[t]
\centering
\caption{Quantitative geometry comparison on the Melbourne test set. Best results are shown in \textbf{bold}.}
\label{tab:geo_mel}
\begin{tabular}{lccccc}
\toprule
Method & CD $\downarrow$ & F-Score $\uparrow$ & JSD $\downarrow$ & COV-CD $\uparrow$ & ~~1-NNA-CD ($\rightarrow$50\%) \\
\midrule
\textbf{Ours}      & \textbf{0.00272} & \textbf{0.8840} & \textbf{0.009752} & \textbf{51.8\%} & \textbf{46.5\%} \\
PVD       & 0.00382 & 0.7939 & 0.036975 & 24.4\% & 69.8\% \\
SemCity   & 0.00451 & 0.7800 & 0.144380 & 4.1\%  & 96.8\% \\
DPM       & 0.01936 & 0.5692 & 0.279454 & 5.2\%  & 98.7\% \\
PointFlow~~ & 0.12071 & 0.2513 & 0.786199 & 0.2\%  & 100.0\% \\
\bottomrule
\end{tabular}
\end{table}

Table~\ref{tab:geo_mel} and Fig.~\ref{fig:geo_compare} present the quantitative and qualitative geometry results on the Melbourne test set. \ours{} achieves the best performance across all five metrics, reducing Chamfer Distance by 29\% relative to the next-best method (PVD) and reaching an F-Score of 0.884. The distributional metrics further highlight the gap: our JSD (0.010) is nearly $4{\times}$ lower than PVD (0.037), and our COV-CD of 51.8\% indicates that our generated tiles cover more than half of the test distribution, compared to 24.4\% for PVD. The 1-NNA-CD of 46.5\%, close to the ideal 50\%, confirms that \ours{} samples are nearly indistinguishable from real tiles in the metric space. PointFlow and DPM
fail to capture the structural complexity of city tiles, producing amorphous geometry. SemCity generates recognizable urban layouts via triplane diffusion but introduces coarse voxelization artifacts and lacks fine-grained building detail.

\subsubsection{Visual Comparison.}

\begin{figure}[t]
    \centering
    \includegraphics[width=1\linewidth]{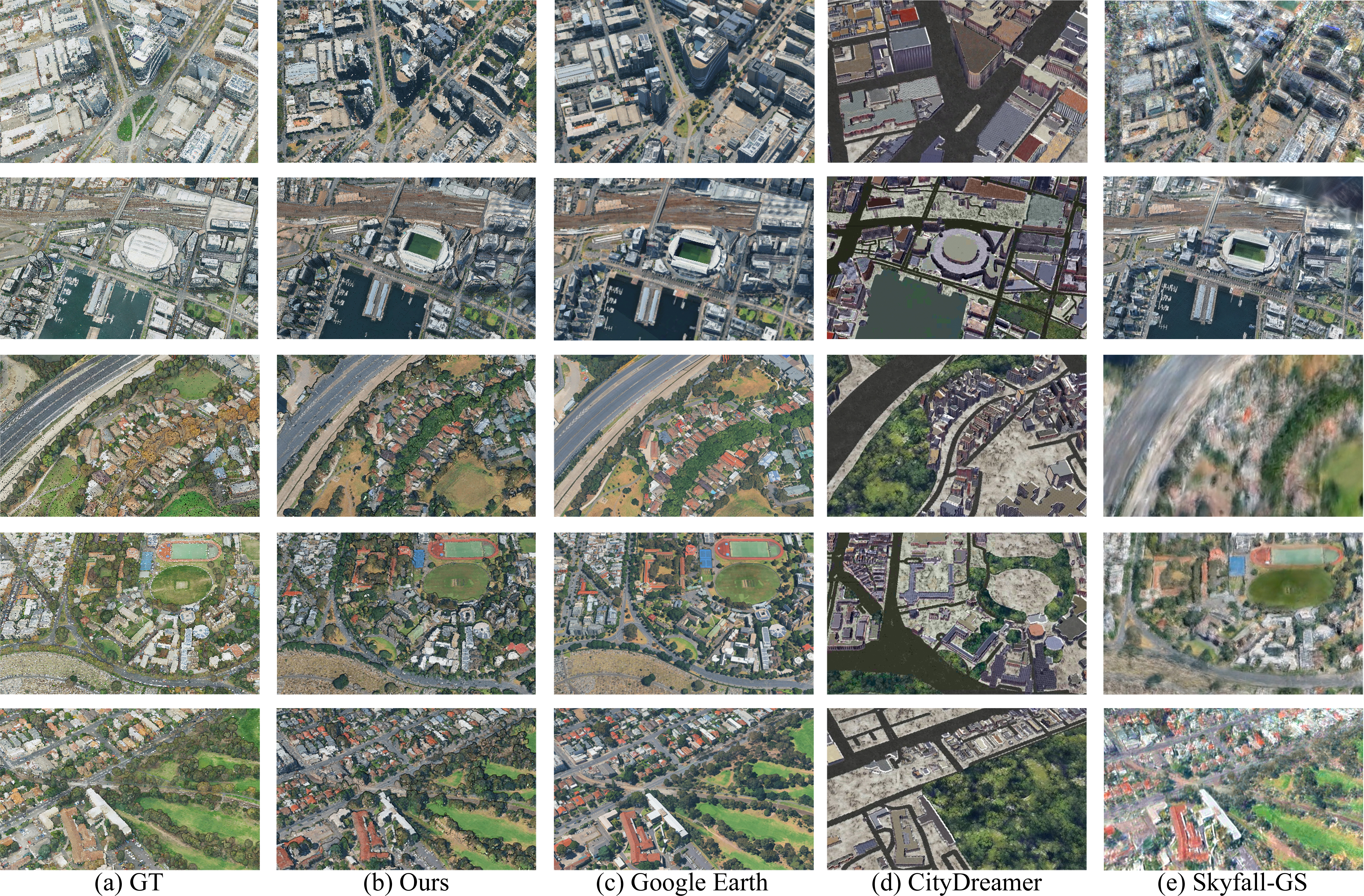}
    \caption{Visual comparison of rendered outputs against Google Earth ground truth. \ours{} produces scene-specific geometry and appearance closest to the reference, while CityDreamer generates city-like but location-agnostic content, and Skyfall-GS produces blurred reconstructions.}
    \label{fig:visual_compare}
\end{figure}

\begin{table}[t]
\centering
\caption{Rendered image quality compared against the conditioning satellite image. Results reflect conditioning fidelity; the satellite-conditioned method has a structural advantage over methods using independent inputs.}
\label{tab:visual}
\begin{tabular}{lccccc}
\toprule
Method & PSNR $\uparrow$ ~~& SSIM $\uparrow$ ~~& LPIPS $\downarrow$ ~~& FID $\downarrow$ ~~& Hist-Corr $\uparrow$ \\
\midrule
Ours          & 22.82 & \textbf{0.70} & 0.30 & \textbf{58.52} & 0.88 \\
Google Earth  & \textbf{26.11} & 0.67 & \textbf{0.28} & 62.83 & \textbf{0.90} \\
CityDreamer   & 13.21 & 0.26 & 0.62 & 155.27 & 0.38 \\
Skyfall-GS   & 16.50 & 0.38 & 0.52 & 115.68 & 0.58 \\
\bottomrule
\end{tabular}
\end{table}

To evaluate the full pipeline including colorization, we render top-down views of generated point clouds and compare against satellite imagery from the same locations (Table~\ref{tab:visual}, Fig.~\ref{fig:visual_compare}). \ours{} achieves an SSIM of 0.70 and FID of 58.52. That \ours{} slightly exceeds Google Earth on SSIM and FID is expected: our method is conditioned on the evaluation satellite image, whereas Google Earth renders come from independent multi-view capture under different lighting. We therefore include Google Earth as a reference point, not a baseline to surpass. CityDreamer produces plausible but location-agnostic imagery (PSNR 13.21, SSIM 0.26), and Skyfall-GS yields blurred outputs (PSNR 16.50, LPIPS 0.52) due to the limited satellite baseline.

\subsubsection{Large-Tile Visualization.}

\begin{figure}[t]
    \centering
    \includegraphics[width=1\linewidth]{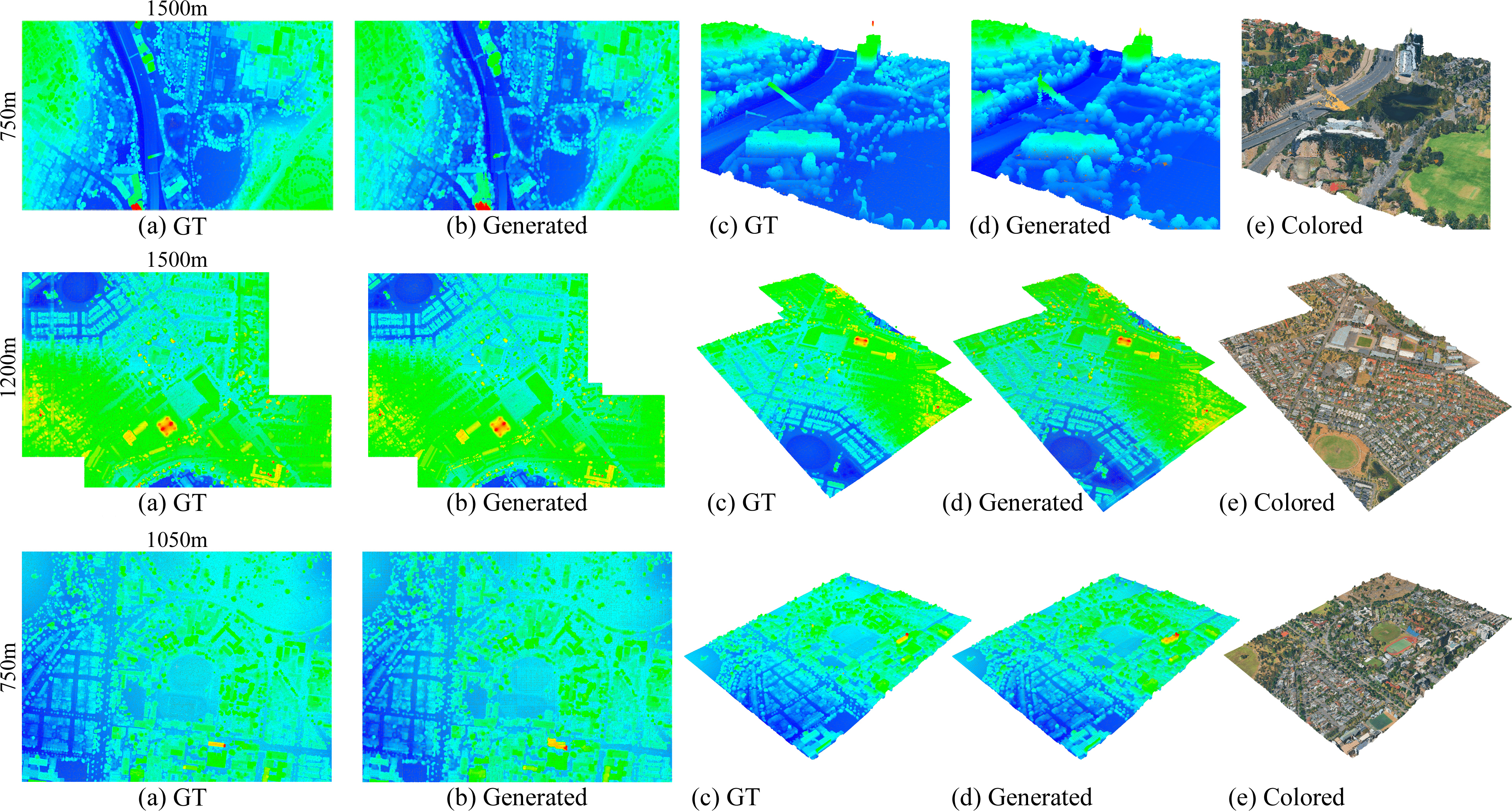}
    \caption{Large-scale visualization of seamlessly tiled Melbourne point clouds generated by \ours{}. Adjacent tiles are independently generated with latent-space edge consistency, producing continuous geometry across tile boundaries.}
    \label{fig:mel_vis}
\end{figure}

Fig.~\ref{fig:mel_vis} shows multiple independently generated tiles assembled from the Melbourne test set. The latent-space edge consistency mechanism (Sec.~\ref{sec:edge}) enables seamless tiling: buildings, roads, and vegetation continue naturally across tile boundaries without visible seams, demonstrating practical viability for city-scale deployment.

\subsubsection{Cross-Dataset Generalization.}

\begin{figure}[t]
    \centering
    \includegraphics[width=1\linewidth]{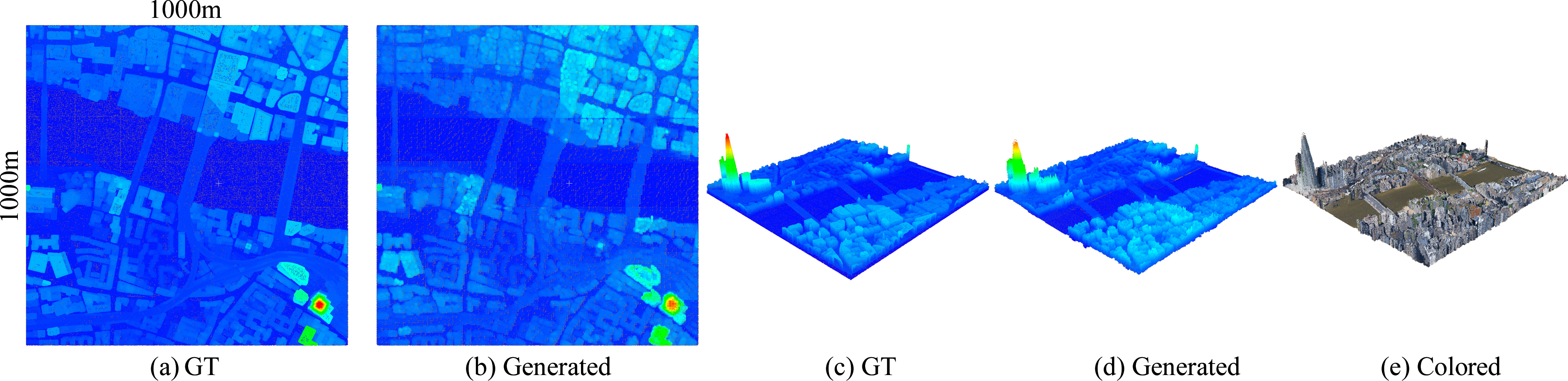}
    \caption{Qualitative visualization on the HoliCity (London) dataset. The model, trained on Melbourne, generalizes to the structurally distinct London cityscape.}
    \label{fig:holi_vis}
\end{figure}

\begin{table}[t]
\centering
\caption{Evaluating \ours{} with other methods, trained on the Melbourne dataset, on the HoliCity (London) test split without retraining or finetuning.}
\label{tab:holi}
\begin{tabular}{lccccc}
\toprule
Method & CD $\downarrow$ & F-Score $\uparrow$ & JSD $\downarrow$ & COV-CD $\uparrow$ & 1-NNA-CD ($\rightarrow$50\%) \\
\midrule
\textbf{Ours} & \textbf{0.00394} & \textbf{0.7826} & \textbf{0.0149} & \textbf{41.8\%} & \textbf{63.7\%} \\
PVD & 0.00542 & 0.7113 & 0.0526 & 23.5\% & 78.9\% \\
SemCity & 0.00613 & 0.6942 & 0.1681 & 6.8\% & 94.5\% \\
DPM & 0.02174 & 0.5221 & 0.3017 & 5.4\% & 97.8\% \\
PointFlow & 0.11261 & 0.2394 & 0.8125 & 0.3\% & 100.0\% \\
\bottomrule
\end{tabular}
\end{table}

To assess generalization beyond the training city, we evaluate the \ours{} trained on Melbourne data (as is without retraining) on the HoliCity~\cite{zhou2020holicity} London test split. As shown in Table~\ref{tab:holi} and Fig.~\ref{fig:holi_vis}, the model achieves a CD of 0.00394 and F-Score of 0.783, degrading only moderately from the Melbourne results despite the substantial domain shift: London features denser, taller buildings with more complex structural layouts compared to Melbourne's largely suburban environment. The COV-CD of 41.8\% and 1-NNA-CD of 63.7\% show that, while not fully capturing London's structural diversity, the model produces plausible outputs that reflect the conditioning inputs, confirming cross-city generalization of the structured latent representation.

\subsubsection{Ablation: Conditioning Modalities.}

\begin{table}[t]
\centering
\caption{Ablation on conditioning modalities. VAE Recon encodes and decodes ground-truth tiles without the flow model, serving as an upper bound. Best generative results in \textbf{bold}.}
\label{tab:ablation_cond}
\begin{tabular}{llcccc}
\toprule
Variant & CD $\downarrow$ & F-Score$\uparrow$ & JSD $\downarrow$ & COV-CD $\uparrow$ & 1-NNA-CD ($\rightarrow$50\%) \\
\midrule
VAE Recon (upper-bnd) & 0.00131 & 0.9861 & 0.0033 & 85.8\% & 47.2\% \\
\midrule
\textbf{Full model}     & \textbf{0.00272} & \textbf{0.8840} & \textbf{0.0098} & 51.8\% & \textbf{46.5\%} \\
w/o Semantic            & 0.00350 & 0.8477 & 0.0340 & \textbf{59.2\%} & 75.1\% \\
w/o DSM                 & 0.00340 & 0.8490 & 0.0206 & 23.0\% & 82.2\% \\
Satellite only          & 0.00389 & 0.8244 & 0.0185 & 23.0\% & 83.9\% \\
w/o CFG                 & 0.00362 & 0.8300 & 0.0108 & 41.0\% & 76.0\% \\
\bottomrule
\end{tabular}
\end{table}

We ablate the contribution of each input modality and classifier-free guidance (CFG) in Table~\ref{tab:ablation_cond}. All variants use the same checkpoint; structured condition dropout ($p{=}0.1$) during training makes inference-time removal methodologically consistent.
Every modality contributes: removing any single condition degrades CD by 25--43\%, with DSM being especially critical for distributional coverage (COV-CD drops to 23.0\% without it). Disabling CFG degrades both fidelity and diversity, confirming its role in sharpening conditional generation. The VAE Recon upper bound shows that the latent space preserves substantial geometric detail, with the remaining gap attributable to the flow-based generation process.

\section{Conclusion}
\label{sec:conclusion}

We presented \ours{}, a multi-stage framework for generating dense, colored city-scale point clouds from satellite imagery and auxiliary remote sensing modalities. The key insight is a Grid-Aligned VAE that compresses $10^5$-point city tiles into a topology-preserving latent grid, enabling spatially coherent multi-modal conditioning and an efficient latent-space edge consistency mechanism for seamless cross-tile generation. A conditional rectified flow model synthesizes geometry latents, and an orientation-aware diffusion colorizer handles the observability asymmetry between horizontal and vertical surfaces. We also introduced City3D-MultiGen, a benchmark of 163K tiles from Melbourne and London with aligned point clouds, satellite images, semantic maps, and elevation data. Experiments demonstrate that \ours{} outperforms existing point cloud generation baselines across all geometry metrics and produces visually coherent colored outputs with seamless boundaries over arbitrarily large urban extents.

\subsubsection{Limitations and Future Work.}
The sequential three-stage training may accumulate errors across stages; joint fine-tuning could mitigate this.
Fa\c{c}ade colorization relies on learned priors since satellite imagery provides no vertical supervision; incorporating street-level images could improve quality.
Cross-city generalization degrades moderately (CD from 0.00272 to 0.00394 on London), and the fixed tile size and cell resolution may not optimally serve all urban morphologies.
Future directions include temporal generation for urban change simulation, multi-resolution latent grids, and scaling the benchmark to additional cities.

\section*{Acknowledgements}
This research was supported by the Australian Research Council Discovery Project DP240101926.

\bibliographystyle{splncs04}
\bibliography{references}

\end{document}